# On the role of autocorrelations in texts

*D.V. Lande, A.A. Snarskii*

The task of finding a criterion allowing to distinguish a text from an arbitrary set of words is rather relevant in itself, for instance, in the aspect of development of means for internet-content indexing [1] or separating signals and noise in communication channels [2].

The Zipf law is currently considered to be the most reliable criterion of this kind [3]. At any rate, conventional stochastic word sets do not meet this law.

The present paper deals with one of possible criteria based on the determination of the degree of data compression.

The most natural approach to solving the above problem is, no doubt, a study of autocorrelations in sequences of words forming a document. Data compression used in various file compression systems, in particular, gzip archiver, can be proposed as a mechanism for study of similar autocorrelations. Really, the compression method it employs is based on coding repeated patterns [4], which, undeniably, in case of text compression, depends, among other things, on the order of words: the more stable patterns are there in the data set, the more efficient is the compression.

It can be supposed that changing the order of words arrangement in a coherent text and destructing thereby its inherent patterns will reduce the level of compression of the result obtained. With full "intermixing" of words in the initially coherent text one might expect a certain minimum value of compression level which no longer depends (or, at least, weakly depends) on further permutation of words.

Thus, studying the degree of data compression for different states of original document obtained by consecutive permutation of words, one can hypothetically construct a quantitative measure of a degree of document "connectivity" which might be naturally considered one of the fundamental text properties.

To check this hypothesis, the authors have studied a variety of well-known literary works. The results obtained in the framework of this study, though call for further investigation, testify in favour of principal applicability of this method.

In the course of the investigation a procedure of words intermixing was used. To determine the intermixing level, special rules were introduced. Let $N$ be a number of words in a text following each other. Under one "atomic" operation of words permutation the following procedure is meant: a generator of preudorandom numbers is used to determine two numbers $n$ and $m$ in the range of values $[1, N]$. Then the words with ordinal numbers $n$ and $m$ change their places.

In the course of the investigation the number of consecutive "atomic" permutations of words was $[kN/10]$ ($k = 0, ..., 20$; $k = 0$ corresponds to the initial state). Thus, 21 states of the original document were obtained and registered. The document in each of the states obtained was subject to compression by means of the Lempel-Ziv algorithm [4], following which its volume corresponding to given state was determined. As a result, dependences of the volumes of compressed documents on the degree of intermixing of words were obtained. Typical appearance of such dependences is given in Fig. 1. Note that this operation was applied to texts of different length.

It can be seen that within $k < 6$ the curve demonstrates a monotonous growth, and then reaches saturation with slight (and evidently random) fluctuations about some average value. Their amplitude makes about 0.1 of full difference between the maximum and minimum compression levels. Taking into consideration that a compressed document on the average is of the order of 30% of the original one, the result we obtained should be regarded as explicit one.

In Fig. 2 are shown the values of $\chi$ (the ratio of established average value of the volume of compressed intermixed text to the volume of compressed original text) as a function of the length of compressed text fragment for different texts – the works by different authors.

As can be seen, the values of $\chi$ grow with increasing volume of texts, achieving certain level that can apparently testify to the authors' styles.

Research was made into a behaviour of parameters $\chi$ for texts of messages of electronic mass media. A body of messages of electronic mass media of volume 5000 documents was used as an experimental basis. The average length of the documents in the body was considerably less than the above considered literary works and made as few as 3951 symbols. Fig. 3 shows a plot of $\chi$ (ordinate axis) versus the number of documents ranked according to this parameter (abscissa axis). It turned out that criterion $\chi > 1$ is met by 4909 documents making 98.18%. It is noteworthy that abnormally low $\chi$ are inherent in the documents with the smallest length (the average length of 91 documents with $\chi \leq 1$ is as small as 834 symbols).

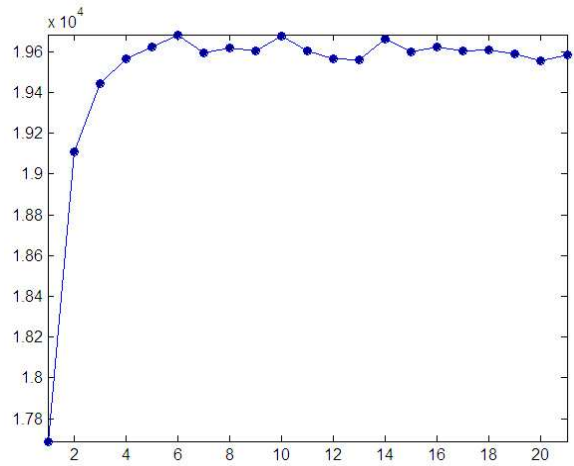

*a)*

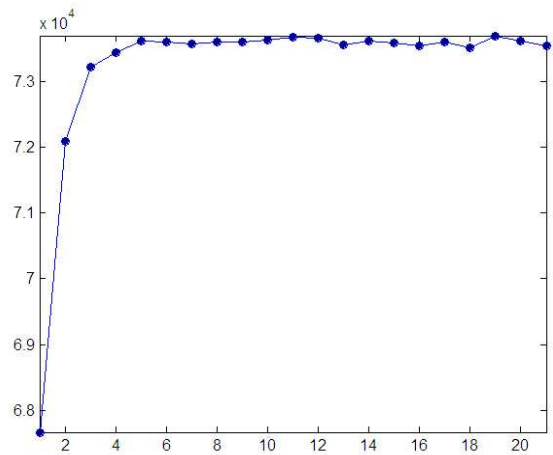

*b)*

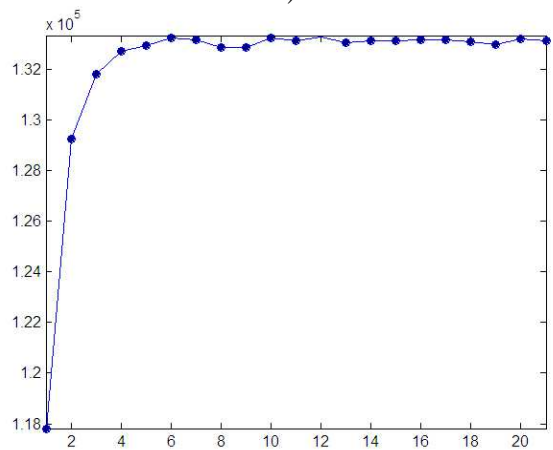

*c)*

*Fig.1. Archive volumes (ordinate axis) versus intermixing coefficient: a) – Richard Bach. Jonathan Livingston Seagull (53149 symbols); b) – William Shakespeare. Hamlet (207402 symbols) ; c) – Ernest Hemingway. Green hills of Africa (363513 symbols)*

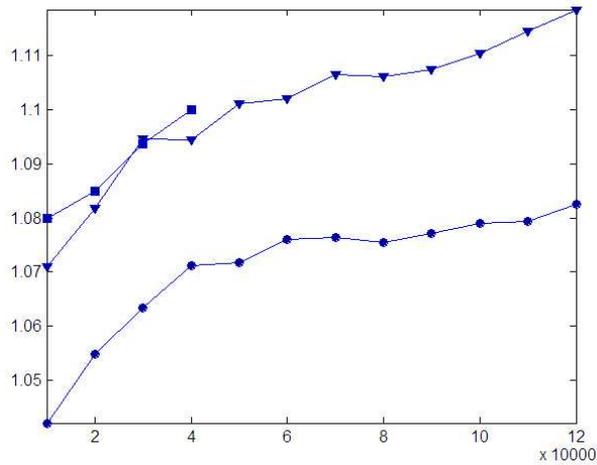

*Fig. 2. The values of $\chi$ versus fragment length:
a) – Richard Bach. Jonathan Livingston Seagull; b) – William Shakespeare. Hamlet ;
c) – Ernest Hemingway. Green hills of Africa*

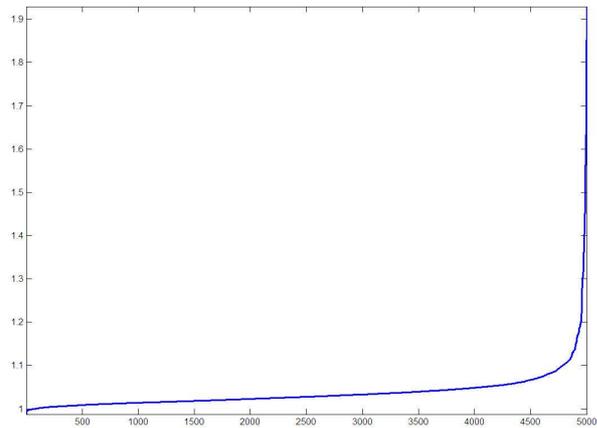

*Fig.3. Rank distributeon of $\chi$*

The authors have also performed an experiment on artificial text generation. For this purpose "artificial" words comprising from 1 to 12 letters were generated. These words were used to form texts about 10000 symbols long. In so doing, individual words in the texts formed were repeated with frequencies that allowed simulation of ratios established by Zipf. Fig. 4 shows the result of comparing the arrays of parameter $\chi$ values for real and artificial texts. The numbers of documents under study are plotted on the abscissa.

In the course of investigation the authors managed to reveal a formalized distinction between random word sets and real texts. The resulting dependence of the volume of compressed text on the degree of intermixing proved to be one of the criteria for determination whether an array of words is a text or a simple word set, even though it formally meets the Zipf law.

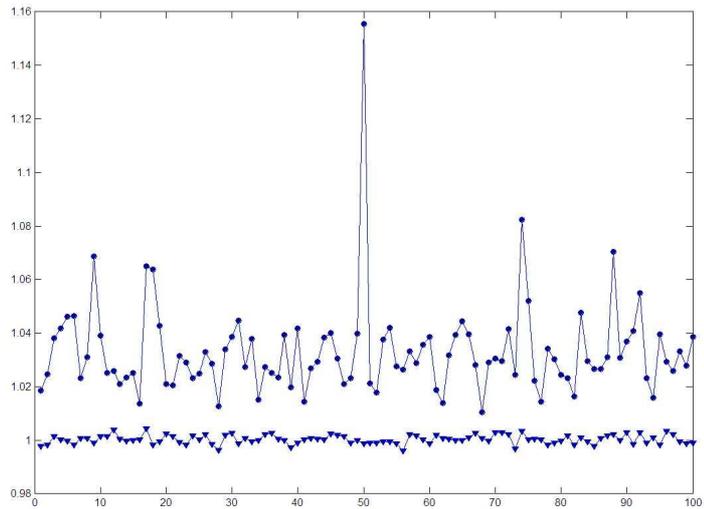

*Fig. 4. Parameter χ for real (●) and artificial (▼) texts*